Oral                                            Topic: *Robotics (Human-Robot Interaction)*

# Human-Robot Interaction: Applications

**Abdel-Nasser Sharkawy**
Mechatronics Engineering, Mechanical Engineering Department
Faculty of Engineering, South Valley University
Qena 83523, Egypt
E-mail: eng.abdelnassersharkawy@gmail.com

**Summary:** Recently, human-robot interaction (HRI) is an extensive research topic and theme which gained importance and significance. HRI aims at the complementary combination between the robot capabilities and human skills. The robots assist humans in terms of precision, speed, and force. The humans contribute in terms of the experience, knowledge of executing the task, intuition, and easy adaptation and learning, and understanding of control strategies. In this work, the applications of human-robot interaction are reviewed. These applications could be industrial, medical, agricultural, servical, and educational. HRI can be found in industrial applications in picking and placing in the production lines, welding processes, parts assembly, and painting. Assistive robotics are one from the highest profile areas in HRI. For people with the physical and the mental challenges, the robots can provide the opportunity of interaction and therapy. In addition, HRI can be widely applied in hospitals. Nowadays, HRI is very important for facing the new coronavirus (COVID-19) pandemic. In agriculture, the cooperation between human and the robot helps with many tasks including harvesting, seeding, fertilizing, spraying, weed detection, hauling, and mowing. HRI can also be found in other applications such as education, mining, and home use.

**Keywords:** Human-Robot Interaction, Industrial Applications, Medical and Rehabilitation Applications, Precision Agriculture.

## 1. Introduction

HRI is the field of the study referring to understanding, designing, and evaluating the robotic systems used by or collaborating with the human operator. Interaction needs the communication between the robots and the humans. The human communication with the robot could take many forms. However, these forms are highly affected by whether the human operator closeness to the robot. Therefore, the communication or interaction between both the human and the robot is separated into two main categories [1], [2]: The first category is the remote interaction and the second is poximate interaction. In the remote one, the human does not exist in nearby the robot. In addition, they are spatially or even temporarily separated. An example for this category is the Mars Rovers which is separated from earth in space and time as well. In the proximate category, the human and robot coexist in the same location. An example for this category is the service robots during their sharing with the humans in the same room.

These categories can help distinguish between the applications requiring the mobility, physical interaction, or social interaction. Remote interaction using mobile robots refers to the tele-operation or the supervised control. Remote interaction using the physical manipulator refers to the tele-manipulation. Proximate interaction using the mobile robots takes the form of the robot assistant. Proximate interaction includes the physical interaction. In social interaction, the social and the emotive as well as the cognitive interaction aspects are included. In social interaction, the humans and the robots interact using the form of peers or companions. Importantly, social interactions with the robots take the form of proximate interaction rather than remote interaction.

From these categories, this manuscript addresses the proximate interaction. It presents some real applications for HRI in industrial environment, medical and rehabilitation, agriculture, education, and other environments. These applications show the importance of the interaction happening between the robot and human operator in our real life.

The rest of this paper is divided as follows. section 2 presents the industrial applications of HRI. In section 3, the HRI in the medical and rehabilitation applications is presented. HRI in agriculture is presented in section 4, whereas section 5 illustrates some other applications. Section 6 concludes the main important point of this manuscript.

## 2. HRI in Industry

HRI is widely used in industrial applications such as picking and placing in the production lines, welding processes, parts assembly, painting, and so on.

Some examples for HRI in the industrial application are presented from Fig. 1 to Fig. 4. The robot workstation is running in the plant of BMW in South Carolina in which the robot helps human operators to perform the assembly of the final door [3] (see Fig. 1). In the door assembly operation, human operator and robots work together. BMW plant has succeeded to implement and develop the direct human-robot cooperation and interaction in the series production.

Human-robot teams are found also in the flexible production lines [4], [5], as presented in Fig. 2. In this figure, the robotic manipulators and the human operators collaborate in handling of the workpieces. Safety is also is very important in such environments, since the proximity of the human operator to the robot that can lead to potential injuries.

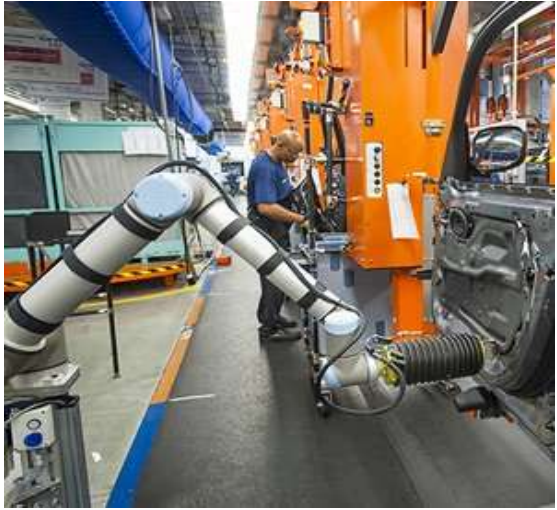

**Fig. 1.** The robot helps the human performing the final door assembly in the BMW plant [3].

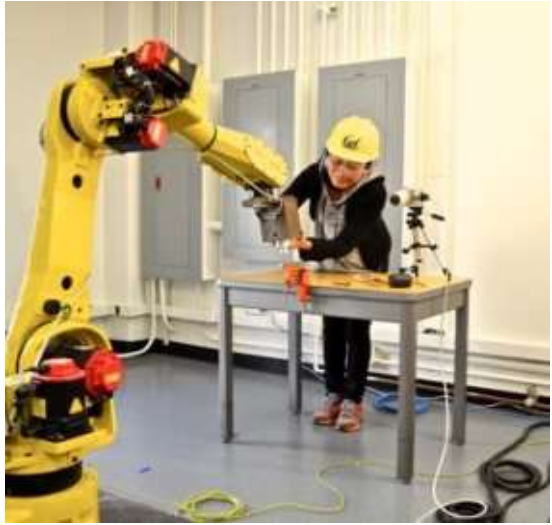

**Fig. 2.** The robot and the human workers cooperate in handling workpieces [4], [5].

Repetitive co-manipulation tasks as shown in Fig. 3 [6] can be performed in suitable poses of the human body. The poses can minimise the effects of the overloading joint torque. Furthermore, they can maximise the capacity of the manipulation of the human.

For handling of heavy and bulky components in welding situations, the multi-robot system with collaborative functionality assists the worker [7], as shown in Fig. 4. Two robots help position the components to be joined in the welding process, at which point the human operator carries out the welding task under favourable ergonomic conditions. In comparison with the standard welding bench, the human operator does not need to assume the uncomfortable postures or the work overhead. All necessary positioning and orientation of the workpieces can be performed by the robot. This also includes presenting of the components in the optimal position for the process of welding, allowing proper flow of the welding bead. Since the robotic repositioning motion is quite fast, the handling time which is about one third of the total process time is reduced to a minimum in comparison to welding processes in which HRI is not implemented.

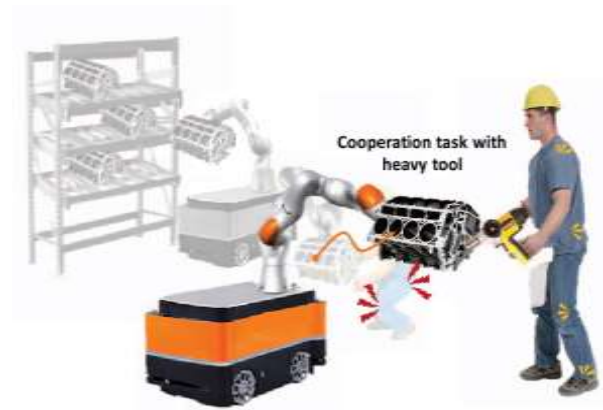

**Fig. 3.** Example for the repetitive co-manipulation tasks [6].

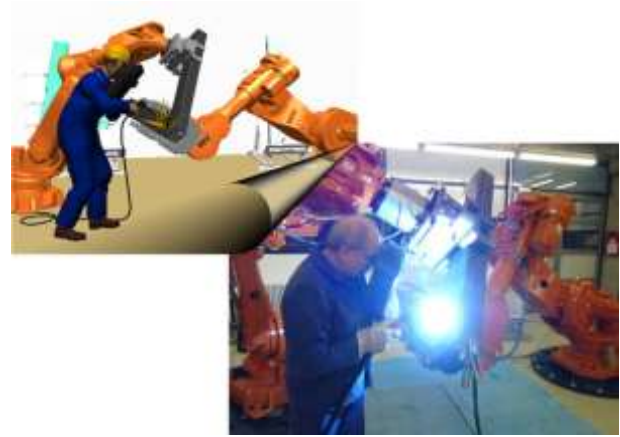

**Fig. 4.** The multi-robot system assists the worker in the welding process [7].

## 3. HRI in Rehabilitation and Medical Applications

Assistive robotics are one of the highest profile areas in HRI. For people with the physical and the mental challenges, robots can provide the opportunity for the interaction and therapy. This work is explored with autistic children in ref. [8], [9]. Many of them cannot respond strongly to the social cues but they respond very well to the mechanical devices. Robots give the possible therapeutic role for using the mechanical device for improving the social interactions [10]. Robots are being considered also for different domains in which children benefit, for example, children with experienced trauma. The social dimension of human-robot interaction is necessary not only in the assistive roles, but also in many areas and domains of the proximate interaction [11]–[15].

For people with physical disabilities, the robot embodiment provides the unique opportunities which are not possible with other technology forms. For example, the researchers are working on the design of the robots which provide and give the support for the physical therapy. Efforts include the providing of the prescribed force and the movement trajectories to help rebuild flexibility as well as strength [16]. Other work for detecting the motivational state and adjusting the therapy in order to maximize the benefits is presented in ref. [17]. Intelligent wheelchairs are the type of the robot which use the external sensors for supporting the path planning and the collision avoidance for the person that requires the wheelchair [18], [19].

An example of the rehabilitation robotics was presented in [20], in which, a robotic system was built as shown in Fig. 5, that performs correctly the exercise for the conventional physiotherapy such as the shoulder flexion, in similar way to what the physiotherapist do during the co-manipulating with KUKA LWR robot. In this way, the excellent characteristics of the human (the capacity of the decision) and the robot (the precision, the work capacity, and the repeatability) could be combined to achieve better results in the musculoskeletal rehabilitation of the arm. Furthermore, the second as well as the main objective was evaluating the non-pathological shoulder behaviour while performing the shoulder flexion movements.

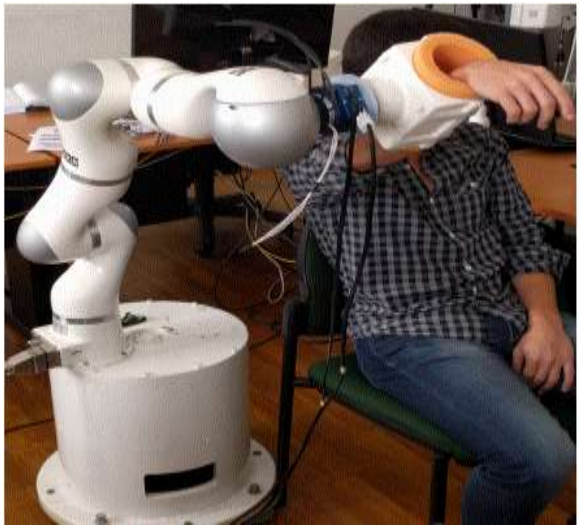

**Fig. 5.** An Example illustrates the rehabilitation robotics [20].

Coronavirus disease (COVID-19) is the infectious disease which is caused by the newly discovered coronavirus. The severity of COVID-19 symptoms can range from mild to severe. The cooperation between the human and the robot help to fight against the new pandemic. Two cases of the human-robot cooperation can be used in the hospital to fight against Coronavirus. The robot can add the solution to the nasopharyngeal swabs from the patients for detecting the coronavirus genetic material [21]. The positive test result illustrates that the patient is infected with the coronavirus. In that case, the samples which are taken from the patients' mouth, nose, and throat are tested for the genetic material of the coronavirus in the laboratory. The staff of the laboratory must only load the samples into the tray and then the COVID-19 test robot takes care with the pipetting. In the second case, a mobile unit is provided to the manipulator. The mobile robot works hand in hand with the human and align to the workpiece with high precision. The mobile robot can be used for providing the patients with the food and medicine. In addition, the robot can measure the temperature of the patients. This would minimize the direct contact between the patients and medical staff and others, and therefore, minimizes the infection potential. The robot can also help the humans for sweeping and washing the floors and the walls. This also minimizes the virus infection.

## 4. HRI in Agriculture

The cooperation between human and robot in agriculture helps with many tasks such as harvesting, seeding, fertilizing, spraying, weed detection, hauling, and mowing.

In the precision agriculture as presented in Fig. 6 [22], the robot helps the human-operator or farmer in picking the strawberry. In that case, safety between the human operator and the robot must be included. Furthermore, the robot was controlled remotely by a co-located operator. Their task was to navigate the robot to the location of pickers when they requested it, allow the filled crates to be loaded onto the robot, and then transport these to the storage facility.In Fig. 7 [23], intelligent robot is installed in the greenhouse to care and help farmers for the melon harvesting.

In Scaffold Mode, the concept of HRI is observed clearly. The human operator and the robot work as a collaborative unified system in which the vehicle autonomously navigates along the structured trees rows whereas the humans on the vehicle concentrate on performing and doing some activities such as 1) thinning, 2) pruning, 3) harvesting, 4) tying trees to wire. Tree trimming tasks were presented with Bergerman et al. [24] and Freitas et al. [25] (see Fig.8) in which the humans working on the robot in the Scaffold Mode were able to trim trees more than twice as fast as humans using the traditional approach based on ladder.

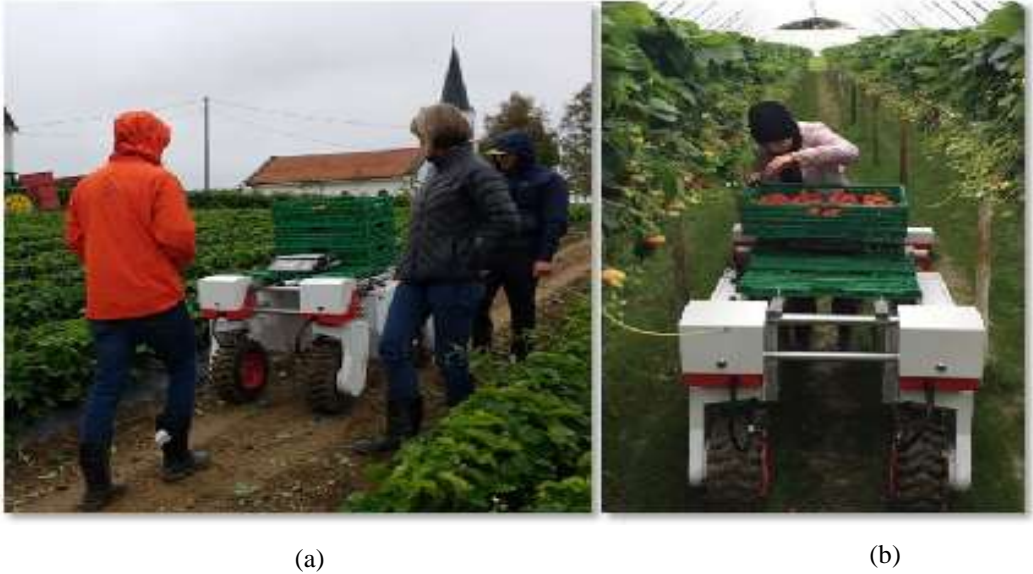

(a)          (b)

**Fig. 6.** The Thorvald robot in the evaluation environment interacting with the pickers: (a) in the open strawberry fields, (b) in the polytunnel environment [22].

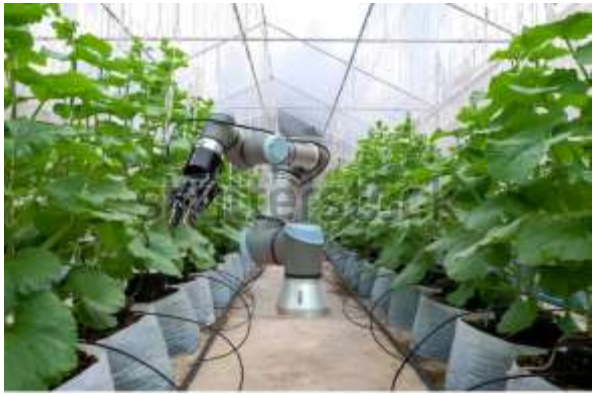

**Fig. 7.** The robot is installed in the greenhouse for helping the farmers for harvesting the melon [23].

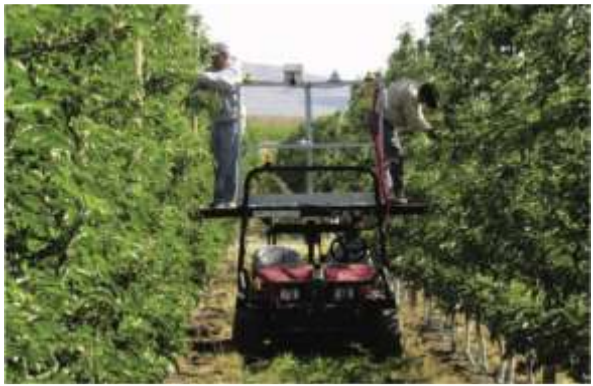

**Fig. 8.** Human operators performing tree trimming while standing in a robot [24], [25].

## 5. Other Applications

HRI can be found in other applications, which include service, home use, inventory management, and mining. In addition, robots is used for promoting the education for the typical children, whether in the home or in the schools [26], [27]. Space exploration and UAVs are also some applications.

Fig. 9 [28] shows the robot helps the human operator in mining industry. Using the robots increases the arsenal of the tools that help and support the miners to work in more safe and more efficient environment.

Caitlyn Clabaugh and her team [29], [30] developed the socially assistive robot (SAR) tutoring system (Fig. 10) for supporting the educators efforts to teach the number concepts to the preschool children. This system was designed iteratively with the input of the education experts for being developmentally appropriate. The system was investigated and tested in the real-world preschool classroom. The collected data were used for training personalized models of the number concepts learning, leveraging the multimodal data, domain knowledge, and also learning style.

To achieve highly efficient human-robot cooperation for all applications discussed above, safety system should be found and implemented in the robotic system. In addition, adapted controllers should follow the human collaborator intention and the environment changes. Therefore, this can lead to human-friendly robots.

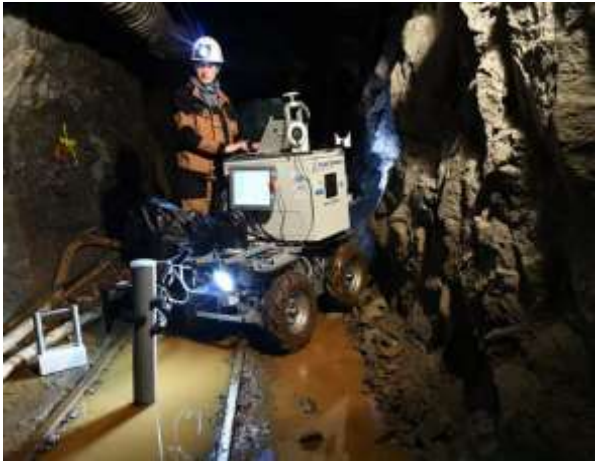

**Fig. 9.** Robot helps human in mining industry [28].

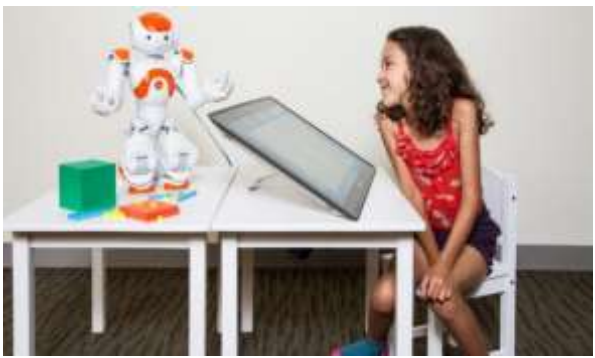

**Fig. 10.** The tutoring system using robot for supporting the educators efforts to teach the number concepts to the preschool children [29], [30].

## 6. Conclusions

This manuscript has summarized some real applications of HRI. HRI could be found in industrial applications in picking and placing in the production lines, welding processes, assembling parts, and painting. Assistive robotics are one of the highest profile areas of HRI. For people with the physical and the mental disabilities, robots provide the opportunity for the interaction and the therapy. In addition, HRI could be commonly be widely applied in hospitals and nowadays HRI is very crucial for fighting against the new coronavirus (COVID-19). In agriculture, HRI helps with many tasks such as harvesting, seeding, fertilizing, spraying, hauling, weed detection, and mowing. HRI can also be found in other applications such as service, home use, inventory management, mining, education, and space exploration.

## Acknowledgements

The authors would like to thank Dr. Ahmed Elsheikh, Assistant Professor, Mechanical Engineering Department, Faculty of Engineering, South Valley University, Qena, Egypt, for revising and checking the English writing of the manuscript.

...